\documentclass{article}

\usepackage[preprint]{corl_2023} 

\usepackage{amsmath}
\usepackage{arydshln}
\usepackage{booktabs}
\usepackage[font=footnotesize]{caption}
\usepackage{enumitem}
\usepackage{float}
\usepackage{graphicx}
\usepackage{multicol}
\usepackage{multirow}
\PassOptionsToPackage{dvipsnames}{xcolor}
\usepackage{xcolor}
\usepackage{wrapfig,lipsum,booktabs}
\hypersetup{
  colorlinks,
  urlcolor  = magenta,
  linkcolor = OrangeRed,
  citecolor = blue,
}

\title{{\huge Giving Robots a Hand:}\\ Learning Generalizable Manipulation with Eye-in-Hand Human Video Demonstrations}

\author{Moo Jin Kim, Jiajun Wu, Chelsea Finn \\
Department of Computer Science \\
Stanford University \\
\small{\texttt{\{moojink,jiajunwu,cbfinn\}@cs.stanford.edu}}
}

\begin{document}
\maketitle

\vspace{-0.8cm}
\begin{figure}[tbh]
    \centering    \includegraphics[width=\linewidth]{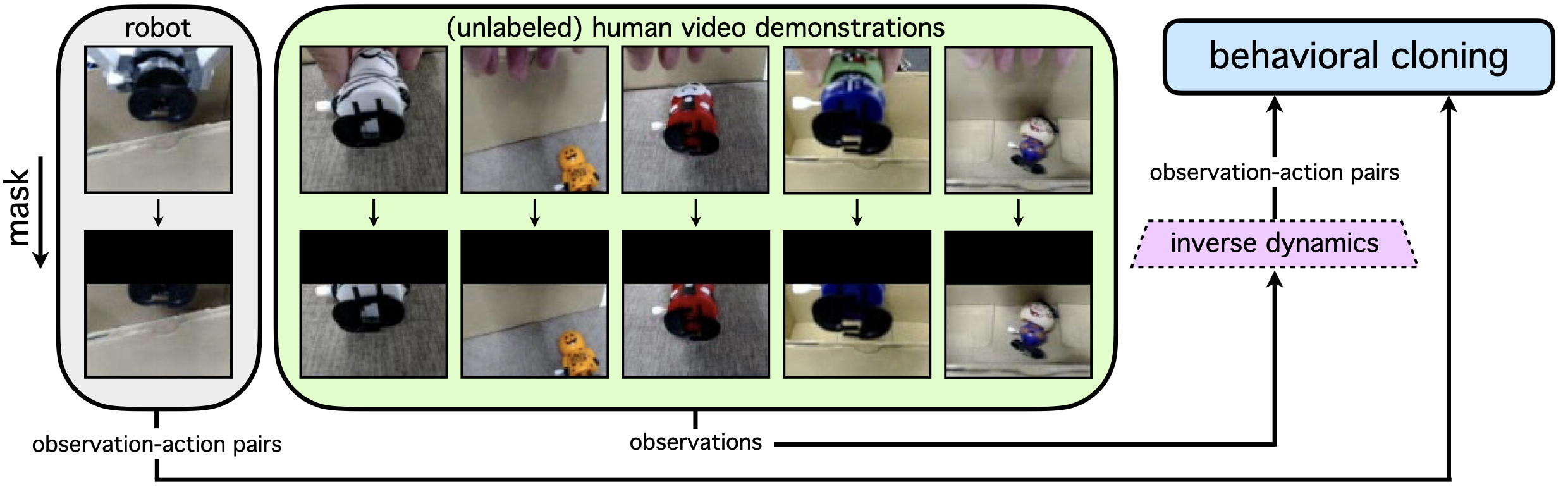}
\caption{We incorporate diverse eye-in-hand human video demonstrations to train behavioral cloning policies that generalize to new environments and new tasks outside the distribution of expert robot imitation data. Images are masked to close the domain gap between human and robot observations. Action labels for human video demonstrations are inferred by an inverse dynamics model trained on robot play data.}
\label{fig:figure1}
\end{figure}

\vspace{-0.2cm}
\begin{abstract}
Eye-in-hand cameras have shown promise in enabling greater sample efficiency and generalization in vision-based robotic manipulation. However, for robotic imitation, it is still expensive to have a human teleoperator collect large amounts of expert demonstrations with a real robot. Videos of humans performing tasks, on the other hand, are much cheaper to collect since they eliminate the need for expertise in robotic teleoperation and can be quickly captured in a wide range of scenarios. Therefore, human video demonstrations are a promising data source for learning generalizable robotic manipulation policies at scale. In this work, we augment narrow robotic imitation datasets with broad unlabeled human video demonstrations to greatly enhance the generalization of eye-in-hand visuomotor policies. Although a clear visual domain gap exists between human and robot data, our framework does not need to employ any explicit domain adaptation method, as we leverage the partial observability of eye-in-hand cameras as well as a simple fixed image masking scheme. On a suite of eight real-world tasks involving both 3-DoF and 6-DoF robot arm control, our method improves the success rates of eye-in-hand manipulation policies by 58\% (absolute) on average, enabling robots to generalize to both new environment configurations and new tasks that are unseen in the robot demonstration data. See video results at \url{https://giving-robots-a-hand.github.io/}.
\end{abstract}


\vspace{-0.3cm}
\section{Introduction}
\vspace{-0.2cm}

Recent works in vision-based robotic manipulation have shown significant performance gains realized by using eye-in-hand cameras in addition to, or in replacement of, static third-person cameras \citep{hsu2022vision, jangir2022look, mandlekar2021matters}. Despite their effects, eye-in-hand cameras alone do not guarantee robust policies, as vision-based models tend to be brittle against real-world variation, such as changes in background, lighting, and object appearances \citep{julian2020never}. Therefore, one natural approach for improving generalization is to train policies on large, diverse robot demonstration datasets \citep{brohan2022rt, jang2022bc, reed2022generalist}. However, collecting such data on a real robot is expensive, as it often requires practitioners to either perform kinesthetic teaching \citep{argall2009survey, billard2006discriminative} or robotic teleoperation \citep{browning2004skill, chen2003programing, mandlekar2018roboturk, ng2004inverted, pook1993recognizing, sweeney2007model, zhang2018deep} via virtual reality headsets or joystick controllers.

In contrast, collecting videos of humans completing tasks is much less expensive because a human operator can rapidly capture many demonstrations without having to constantly reset the robot to some initial state, debug hardware-related issues, or arduously relocate the robot to varied settings to increase visual diversity. Consequently, human video demonstrations are a promising data source that could improve the generalization capabilities of vision-based robotic manipulators at scale.

Despite this enticing potential, a central challenge in learning from human video demonstrations is the difference in appearance between human and robot morphologies, which creates a distribution shift that must be accounted for. Prior works that utilize a \emph{third-person} camera perspective have aimed to mitigate this domain gap by taking explicit domain adaptation approaches, such as performing human-to-robot image translation, learning domain-invariant visual representations, and leveraging keypoint representations of human and robot states (see Section \ref{sec:related-work} for details). In contrast, since we learn policies from an \emph{eye-in-hand} visual perspective, we close the domain gap in a far less involved way: we simply mask a fixed portion of every image such that the human hand or robotic end-effector is no longer visible. As a result, we do not need to employ any domain adaptation method and can learn vision-based manipulation policies end-to-end directly from human videos (where actions are inferred by an inverse dynamics model, which we discuss later). We can thus avoid errors produced by explicit domain adaptation methods, e.g., conspicuous visual artifacts from human-to-robot image translations \citep{smith2019avid}.

The main contribution of this work is the study of a simple, novel method that incorporates diverse eye-in-hand human video demonstrations to improve environment and task generalization. Across several real-world robotic manipulation tasks, including reaching, grasping, pick-and-place, cube stacking, plate clearing, and toy packing, we observe that our method leads to significant improvements in generalization. Our policies generalize to both new environments and new tasks that are not seen in the robot demonstrations, even in tasks with heavy visual occlusion and multiple stages. On average, we observe a $58$\% improvement in \textit{absolute} success rates across unseen environments and tasks when comparing against policies trained only on robot demonstrations.

\vspace{-0.2cm}
\section{Related Work}
\label{sec:related-work}
\vspace{-0.3cm}

Imitation learning is a powerful paradigm for training an agent to complete a task by learning a mapping between observations and actions. Traditional approaches to robotic imitation assume access to expert demonstrations collected from the robot's observation and action spaces \citep{argall2009survey, atkeson1997robot, hayes1994robot, osa2018algorithmic}. Since collecting expert trajectories with a real robot can be physically demanding or require special teleoperation equipment and training \citep{browning2004skill, chen2003programing, mandlekar2018roboturk, ng2004inverted, pook1993recognizing, sweeney2007model, zhang2018deep}, we study the setting of training robots to complete tasks by watching videos of a human demonstrator. One central challenge here is the distribution shift caused by apparent visual differences between human and robot structures.

Past works have addressed this distribution shift in various ways. Some have employed explicit domain adaptation techniques such as human-to-robot context translation \citep{liu2018imitation, sharma2019third} and pixel-level image translation \citep{smith2019avid, li2021meta, xiong2021learning}, commonly using generative models like CycleGAN, which can learn mappings between domains given unpaired data \citep{zhu2017unpaired}. Other works have explicitly specified the correspondences between human and robot embodiments and behaviors by, e.g., employing pose and object detection techniques \citep{bahl2022human, kumar2022graph, lee2022learning, nguyen2018translating, ramirez2017transferring, yang2015robot, simeonov2022neural, wen2022you} and learning keypoint-based state representations of human and robot observations \citep{das2020model, xiong2021learning}. Some have taken a more implicit approach and learned domain-invariant visual representations or reward functions that are useful for solving downstream tasks \citep{sermanet2016unsupervised, sermanet2018time, yu2018one, yang2019learning, mees2020adversarial, schmeckpeper2020reinforcement, chen2021learning, zhou2021manipulator, alakuijala2022learning, nair2022r3m, zakka2022xirl}. Yet another class of works used robotic end-effectors more closely resembling the human hand (e.g., Allegro Hand) to train dexterous manipulation policies via hand pose estimation and kinematic retargeting \citep{handa2020dexpilot, qin2021dexmv, arunachalam2022dexterous, qin2022one, sivakumar2022robotic}.

In contrast to most of these works, which use human demonstrations captured from \emph{third-person} cameras, we avoid the need to apply any explicit domain adaptation or human-robot correspondence mapping method by utilizing \emph{masked eye-in-hand} visual inputs. We also train policies that generalize to new settings and tasks without having to learn intermediate representations or reward functions. Further, unlike prior works utilizing manipulators with more than two fingers, we employ a parallel-jaw robotic end-effector despite it being visually and kinematically dissimilar from the human hand.

Relatedly, \citet{young2020visual} and \citet{song2020grasping} amass diverse manipulation data using ``reacher-grabber'' tools. To minimize domain shift, these tools are attached to the robot arms or engineered to closely resemble real parallel-jaw end-effectors. In contrast, we collect demonstrations with the human hand, which is faster and more flexible than these tools, and test our policies directly on a robot with a structurally dissimilar gripper. Further, our lightweight eye-in-hand camera configuration for human demonstrations is simple to assemble and has nearly zero cost (aside from purchasing the camera itself), while the reacher-grabber tool proposed by \citet{song2020grasping} requires more sophisticated assembly and costs approximately \$450 USD (excluding the cost of the camera).

Lastly, the purpose of this work is \emph{not} to perform a head-to-head comparison between eye-in-hand and third-person perspective methods. We defer such discussion to \citet{hsu2022vision} and \citet{jangir2022look}, which already analyze and compare eye-in-hand and third-person methods in robotic manipulation comprehensively. In this paper, we focus specifically on the eye-in-hand visual setting and compare our approach to methods that are compatible to this regime.

\vspace{-0.2cm}
\section{Preliminaries}
\label{sec:preliminaries}
\vspace{-0.3cm}

\textbf{Observation and action spaces.} The observation spaces of the robot and human, $\mathcal{O}^r$ and $\mathcal{O}^h$ respectively, consist of eye-in-hand RGB image observations $o^r \in \mathcal{O}^r, o^h \in \mathcal{O}^h$. The robot's action space $\mathcal{A}^r$ either has four dimensions, consisting of 3-DoF end-effector position control and 1-DoF gripper control, or seven dimensions, consisting of additional 3-DoF end-effector rotation/orientation control. We assume that the human's action space $\mathcal{A}^h$ is the same as the robot's: $\mathcal{A}^h = \mathcal{A}^r$.

\textbf{Problem definition.} Our objective is to incorporate broad human data to train a policy that generalizes better than one that is trained solely on robot data. While broad data can improve generalization along a number of axes, we specifically aim to improve performance in terms of environment generalization and task generalization. We define \textbf{environment generalization} as the ability to execute a learned task in a new environment unseen in the robot demonstrations. We define \textbf{task generalization} as the ability to execute a new, longer-horizon task when the robot demonstrations only perform an easier, shorter-horizon task.

\vspace{-0.2cm}
\section{Learning from Eye-in-Hand Human Video Demonstrations}
\label{sec:methods}
\vspace{-0.2cm}

We now discuss each module of our framework (Figure \ref{fig:figure1}). We first collect eye-in-hand human demonstrations with a simple low-cost setup (Section \ref{sec:data-collection-setup}). We then label human demonstrations with actions using an inverse dynamics model trained on robot ``play'' data (Section \ref{sec:inverse-model}). Finally, we utilize human demonstrations to train generalizable imitation learning policies (Section \ref{sec:bc-with-human-data}).

\subsection{Eye-in-Hand Video Data Collection}
\label{sec:data-collection-setup}
\vspace{-0.2cm}

\begin{figure}[tb]
    \centering
    \includegraphics[width=0.9\linewidth]{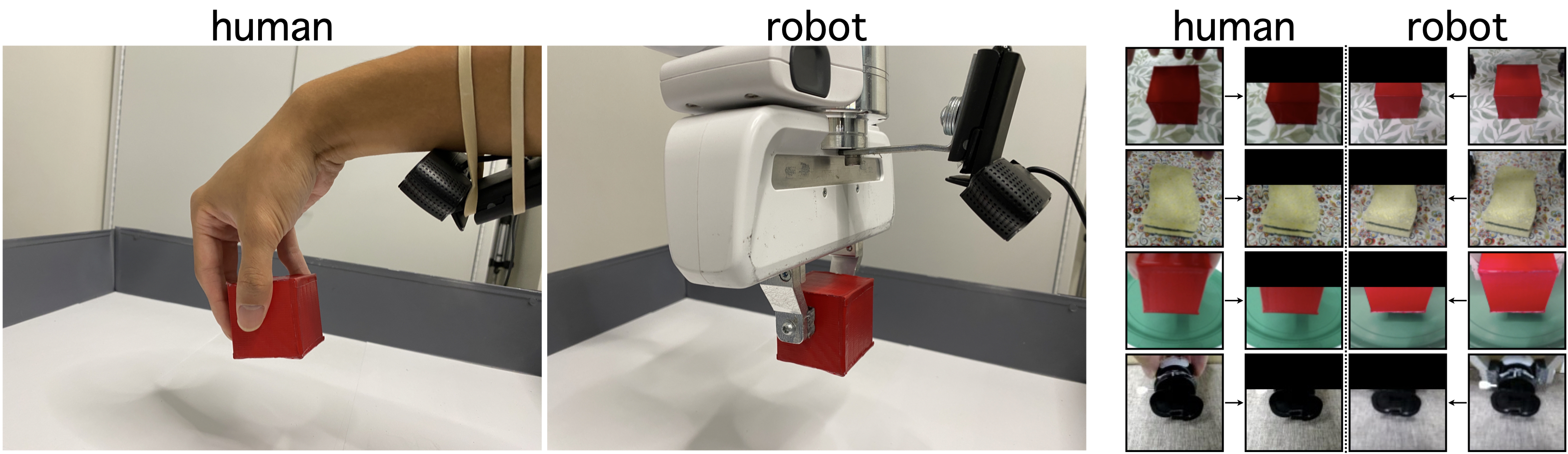}
\caption{\textbf{Left:} Human and robot eye-in-hand camera configurations. Fastening a USB camera on a human arm only requires two rubber bands. Mounting a camera on a Franka Emika Panda robot arm involves L-brackets, washers, and screws. \textbf{Right:} Sample image observations captured by the eye-in-hand human and robot cameras. We mask the top $36\%$ of every image in both domains.}
\label{fig:img-obs-examples-and-wrist-cam-setup}
\vspace{-0.55cm}
\end{figure}

\textbf{Data collection setup.} As shown in Figure~\ref{fig:img-obs-examples-and-wrist-cam-setup}, we secure an RGB camera to a human demonstrator's forearm with two rubber bands, and the demonstrator is immediately ready to collect video demonstrations of a task. While more secure ways of fastening the camera exist, we find that this simple configuration is sufficient and only takes a few seconds to prepare. The same camera is mounted onto a Franka Emika Panda robot arm via an L-bracket assemblage (see Figure \ref{fig:img-obs-examples-and-wrist-cam-setup}). To control the robot, we perform teleoperation with a virtual reality controller (Oculus Quest).

\textbf{Masking the hand and end-effector.} To close the gap between human and robot domains, we mask a fixed region of all image observations $o^h, o^r$ captured by the eye-in-hand human and robot cameras to hide the agent's embodiment. Specifically, we capture images of size $100 \times 100$ and zero out the top $36$ rows of pixels with a script; we denote the resulting human and robot observations as $\bar{o}^h, \bar{o}^r$, respectively. This transformation is shown in Figure \ref{fig:img-obs-examples-and-wrist-cam-setup}. We train inverse dynamics models and imitation learning policies (discussed in subsequent sections) solely on masked images.\footnotemark

\footnotetext{At first glance, it may seem impossible to learn with $\bar{o}^h, \bar{o}^r$ given that the hand or end-effector is not visible. However, we observe that inverse models trained on data in this format can reasonably infer environment dynamics nonetheless due to the presence of certain visual cues. For example, the grasping and lifting of an object can be inferred even when the gripper is not visible due to visual signals such as the object ``locking'' into place as it is secured in the hand, the object beginning to levitate, shadows forming underneath the object, and neighboring objects shrinking in size in the eye-in-hand camera's field of view. Similarly, imitation learning policies can also succeed at various tasks without seeing the hand or end-effector in the frame after a small modification to the policies' inputs (see Section \ref{sec:bc-with-human-data} for details). Nonetheless, masking the image does place some limitations on the tasks that can be performed, which we discuss further in Section~\ref{sec:conclusion}.}

\subsection{Action Labeling of Human Video Demonstrations via Inverse Dynamics}
\label{sec:inverse-model}
\vspace{-0.2cm}

Suppose we have a diverse set of eye-in-hand human video demonstrations for a manipulation task: $\mathcal{D}_\text{exp}^h = \{ \bar{o}_t^h \}_{1...M}$, where $M$ is the total number of timesteps. Since human videos only contain sequences of images, we cannot train an imitation learning policy on this dataset until we generate action labels. The inverse dynamics model serves this precise purpose: Given image observations $\bar{o}_t^h$ and $\bar{o}_{t+1}^h$ at timesteps $t$ and $t+1$, the inverse model predicts the action $a_{t}$ giving rise to the change in observations \citep{nair2017combining, sharma2019third, wang2019learning, schmeckpeper2020reinforcement, li2021meta}. See Appendix \ref{app:inverse-model-arch} for details on the inverse model architecture.

\textbf{Robot play data.}
An inverse model should be trained on data with sufficient diversity such that it can make accurate predictions on diverse human demonstration data. In this paper, we choose to train the inverse model using visually and behaviorally diverse, task-agnostic robot ``play'' data that is collected in a similar manner as \citet{lynch2020learning}. See Appendix \ref{app:play-data-details} for details on how we collect the play data and why it is easy to collect in large quantities.

\textbf{Inverse dynamics model training.} Given robot play data, we now have observation-action transitions $(\bar{o}_{t}^r, a_{t}^r, \bar{o}_{t+1}^r) \in \mathcal{D}_\text{play}^r$. The inverse model, parameterized by $\theta$, takes as input $(\bar{o}_t^r, \bar{o}_{t+1}^r)$ and outputs a prediction $\hat{a}_t^r = f_\theta(\bar{o}_t^r, \bar{o}_{t+1}^r)$. We optimize the parameters $\theta$ to minimize the $L_{1}$ difference between $\hat{a}_t^r$ and $a_t^r$ for $K$ transitions sampled from the play dataset, using Adam optimization \citep{kingma2014adam}:
$ \mathcal{L}(\hat{a}_t^r, a_t^r; \theta)_{1...K} = \sum_{t=1}^{K}  || \hat{a}_t^r - a_t^r ||_1$.

\textbf{Labeling human video demonstrations.}
Once we have trained an inverse model, we run it on all pairs of observations in the human demonstration dataset, $(\bar{o}_t^h, \bar{o}_{t+1}^h) \in \mathcal{D}_\text{exp}^h$, to automatically generate action labels for the demonstrations (see Appendix~\ref{app:inverse-model-analysis} for sample inverse model predictions and analysis). We then have a labeled set of human observation-action pairs, which we denote as $\widehat{\mathcal{D}}_\text{exp}^h = \{ (\bar{o}_t^h, \hat{a}_t^h) \}_{1...M}$, where $M$ is the total number of such pairs. We use this dataset to train an imitation learning policy, as described in the next section.

\vspace{-0.1cm}
\subsection{Imitation Learning with Human and Robot Demonstrations}
\label{sec:bc-with-human-data}
\vspace{-0.2cm}

\textbf{Behavioral cloning.} Given a dataset of human video demonstrations with inferred action labels $\widehat{\mathcal{D}}_\text{exp}^h = \{ (\bar{o}_t^h, \hat{a}_t^h)_{1...M} \}$, we train a manipulation policy via behavioral cloning (BC), which learns a mapping between observations encountered by an expert demonstrator and their corresponding actions \citep{bain1995framework}. In this case, we treat actions $\hat{a}_t^h$ inferred by the inverse model as ``ground truth'' labels representing the demonstrator's actions. The BC policy $\pi_\phi$ takes as input an RGB image observation $\bar{o}_t^h$ and outputs an action $\tilde{a}_t^h$ to best match $\hat{a}_t^h$. We minimize the negative log-likelihood of the predictions to find the optimal policy parameters $\phi^*$, using Adam optimization \citep{kingma2014adam} to train the model.

\textbf{Conditioning the behavioral cloning policy on grasp state.} We modify the BC policy to be conditioned on an additional binary variable $s_t^h$ representing the grasp state at time $t$ (open/closed). This variable provides proprioceptive information about the manipulator that was removed from the image observations by the image masking scheme discussed in Section \ref{sec:data-collection-setup}; without knowing the grasp state, the policy may not be able to discern whether it has already grasped an object and could fail to proceed to complete the task. We automatically estimate $s_t^h$ by setting it as the prior timestep's grasping action, which is inferred by the inverse model when labeling human demonstrations with actions. We then concatenate $s_t^h$ to the latent image embedding and feed the result into the policy network (see Appendix \ref{app:bc-policy-arch} for model architecture details). The resulting policy is $\pi_\phi(\tilde{a}_t^h | \bar{o}_t^h, s_t^h)$, and we optimize $\phi$ as described before.

\textbf{Generalizing beyond narrow robot demonstrations.} As discussed in Section \ref{sec:preliminaries}, we collect and train a BC policy on a narrow set of robot demonstrations and a broader set of human demonstrations with the goal of generalizing to the environments or tasks covered by the human data. The final objective, given $N$ robot samples and $M$ human samples, is to find:
\;\; $ \phi^* = \text{arg min}_\phi - \sum_{t=1}^{N} \log \pi_\phi(\tilde{a}_t^r | \bar{o}_t^r, s_t^r) - \sum_{t=1}^{M} \log \pi_\phi(\tilde{a}_t^h | \bar{o}_t^h, s_t^h)$.

\definecolor{dark-green}{rgb}{.058,.58,.11}
\definecolor{dark-red}{rgb}{.68,.06,.06}
\definecolor{blue}{rgb}{.13,.38,.79}
\definecolor{pink}{rgb}{.89,.07,.64}
\definecolor{yellow}{rgb}{.80,.73,.21}

\vspace{-0.2cm}
\section{Experiments}
\label{sec:experiments}
\vspace{-0.3cm}

We execute a set of experiments to study whether our framework for incorporating broad eye-in-hand human video demonstrations can be used to improve environment generalization and task generalization, as defined in Section \ref{sec:preliminaries}. We then ablate key components of our framework, such as image masking and grasp state conditioning, to study their contributions to the final performance.

\vspace{-0.2cm}
\subsection{Experimental Setup}
\vspace{-0.2cm}

As it is difficult to generate realistic human data in simulation, we perform all experiments in the real world. All observations $o^h \in \mathcal{O}^h, o^r \in \mathcal{O}^r$ are $(3, 100, 100)$ RGB images. Raw image pixels range between $[0, 255]$, but we normalize them to $[-0.5, 0.5]$. We use 6-DoF end-effector control for the toy packing task and 3-DoF control for the rest. In 3-DoF control, the three actions are continuous values ranging between $[-1,1]$ that command the change in the end-effector's position in the Cartesian space. In 6-DoF control, the additional three actions command the end-effector's change in orientation (in extrinsic Euler angles). Lastly, one degree of freedom represents the binary gripper action ($-1$:~close, $1$:~open).

\subsection{Environment Generalization Experiments}
\label{sec:env-gen-experiments}
\vspace{-0.2cm}

\begin{figure*}[t]
    \centering
    \includegraphics[width=0.95\linewidth]{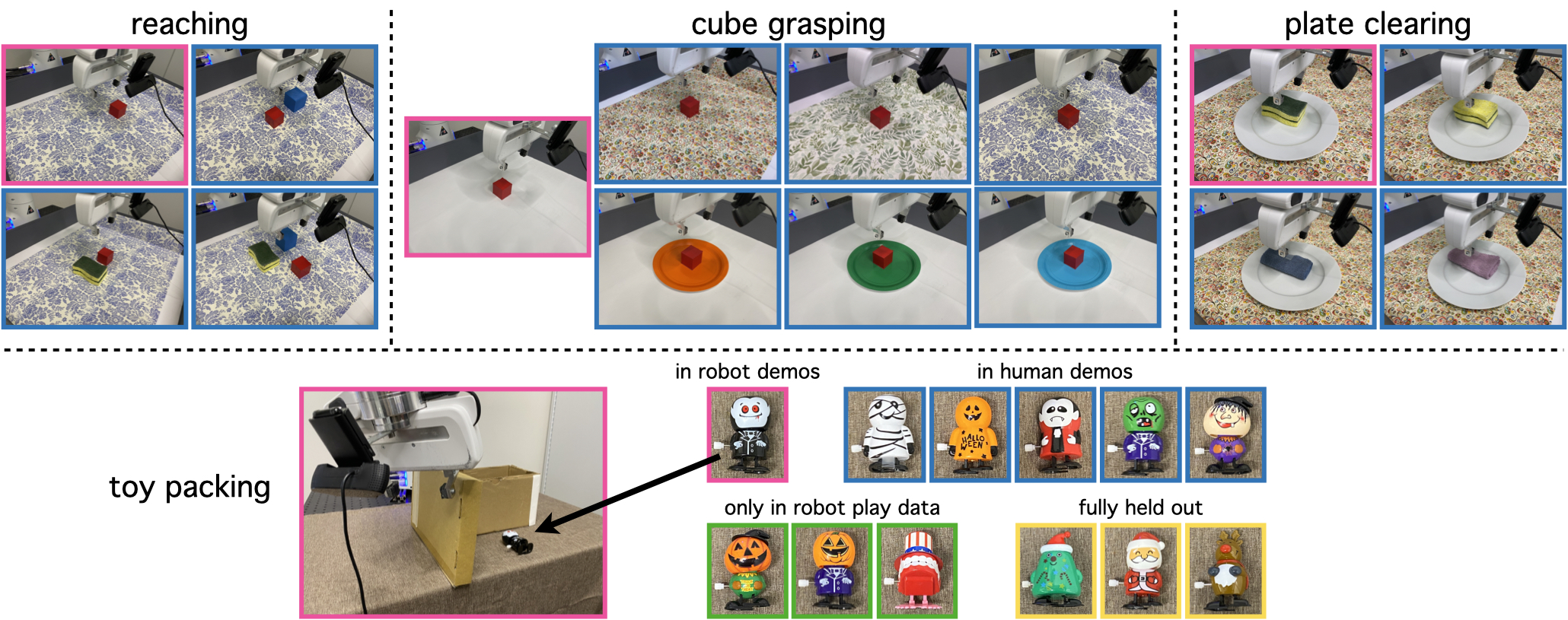}
\caption{Tasks used for environment generalization experiments. Robot demonstrations are collected only in the environment configurations highlighted in {\color{pink}pink}, while human demonstrations are collected in the configurations highlighted in {\color{blue}blue}. In the toy packing task, the toys highlighted in {\color{dark-green}green} are not seen in the human or robot demonstrations but appear in the robot play dataset, while the toys highlighted in {\color{yellow}yellow} are not seen at all in human or robot play/demonstration data (i.e., these are fully held out).}
\label{fig:env-gen-tasks}
\vspace{-0.4cm}
\end{figure*}

Recall that environment generalization (Section \ref{sec:preliminaries}) is the ability to complete a learned manipulation task in a new environment unseen in the robot demonstration dataset.

\textbf{Tasks.} The tasks include reaching towards a red cube in the presence of different distractor objects, grasping a red cube placed on various environment backgrounds, clearing different objects off of a plate, and packing different toys into a box. See Figure \ref{fig:env-gen-tasks} for a visualization of these tasks and Appendix \ref{app:task-details} for details about each task. The tasks are ordered by increasing difficulty and complexity. The final 6-DoF toy packing task is particularly challenging because it involves heavy occlusion (from a wall positioned between the end-effector and target object); execution of a multi-stage trajectory (reaching around wall, grasping toy, lifting toy, reaching box, dropping toy into box); and three more degrees of freedom than the previous tasks (for end-effector orientation control).

\textbf{Datasets.} For each task, we collect narrow robot demonstrations in one environment, and broad human demonstrations in multiple environments (shown in Figure~\ref{fig:env-gen-tasks}). We also collect a robot play dataset for an inverse model that is shared with a task generalization experiment involving similar objects. See Appendix \ref{app:datasets} for details on all expert demonstration datasets and robot play datasets.

\textbf{Methods.}
In our method, we train a BC policy on robot demonstrations and human demonstrations with the image masking scheme discussed in Section \ref{sec:data-collection-setup}. As we wish to study whether incorporating broad human demonstrations into training achieves increased environment generalization, we compare our method against a baseline policy trained only on narrow robot demonstrations. In addition, to assess whether any improvements in generalization are simply correlated to the increase in training dataset
size, we also compare against a policy trained on both robot demonstrations and robot play data, as the play datasets are larger than the human demonstration datasets. Lastly, we evaluate how effective our image masking method is compared to explicit domain adaptation approaches such as pixel-level image translation by comparing against a policy trained on both human and robot demonstrations, where a CycleGAN is used to translate human images into robot images (as in \citep{smith2019avid} and \citep{li2021meta}). To summarize, we evaluate the following four methods in our experiments, where each one is a BC policy trained on a different set of data:
\vspace{-0.4em}
\begin{itemize}[leftmargin=2.5em]
    \item \textbf{robot}: robot demos only
    \vspace{-0.4em}
    \item \textbf{robot + play}: robot demos and robot play data
    \vspace{-0.4em}
    \item \textbf{robot + human w/ CycleGAN}: robot demos and CycleGAN-translated human demos
    \vspace{-0.4em}
    \item \textbf{robot + human w/ mask (ours)}: robot demos and human demos with image masking
\end{itemize}

\textbf{Results.} As shown in Table \ref{tab:env-gen-and-task-gen-results}, incorporating diverse human video demonstrations into policy training with image masking significantly improves generalization. The policy generalizes to new environment configurations unseen in the robot demonstrations (see fine-grained results in Table~\ref{tab:ind-env-gen-results}). To our knowledge, this marks the first time that a real robot policy is directly trained end-to-end on eye-in-hand human demonstrations. On the other hand, the policy trained only on a limited set of robot demonstrations fails completely in many cases, as shown in Figure~\ref{fig:rollouts}(c), since novel out-of-distribution visual inputs confuse the policy. In addition, we see that a policy also trained on the full play dataset, which is larger than the set of human demonstrations, does not perform as well as one trained on the human demonstrations, verifying that generalization performance is not simply a function of training dataset size. Further, while using CycleGAN-translated human demonstrations generally leads to greater performance than using only robot demonstration or play data, it is not as effective as our image masking method. In particular, while the CycleGAN image translations are successful in some cases, they are noisy in other cases (sample translations are shown in Figure~\ref{fig:cyclegan-examples} in Appendix~\ref{app:cyclegan-analysis}); such noise hinders final policy performance. Videos of the policies and extensive qualitative analysis of individual methods are available on our \href{https://giving-robots-a-hand.github.io/}{project website}.

\subsection{Task Generalization Experiments}
\label{sec:task-gen-experiments}
\vspace{-0.2cm}

Recall that task generalization (Section \ref{sec:preliminaries}) is the ability to complete a task that is unseen and longer-horizon than those in the robot demonstrations.

\begin{figure*}[t]
    \centering    \includegraphics[width=0.80\linewidth]{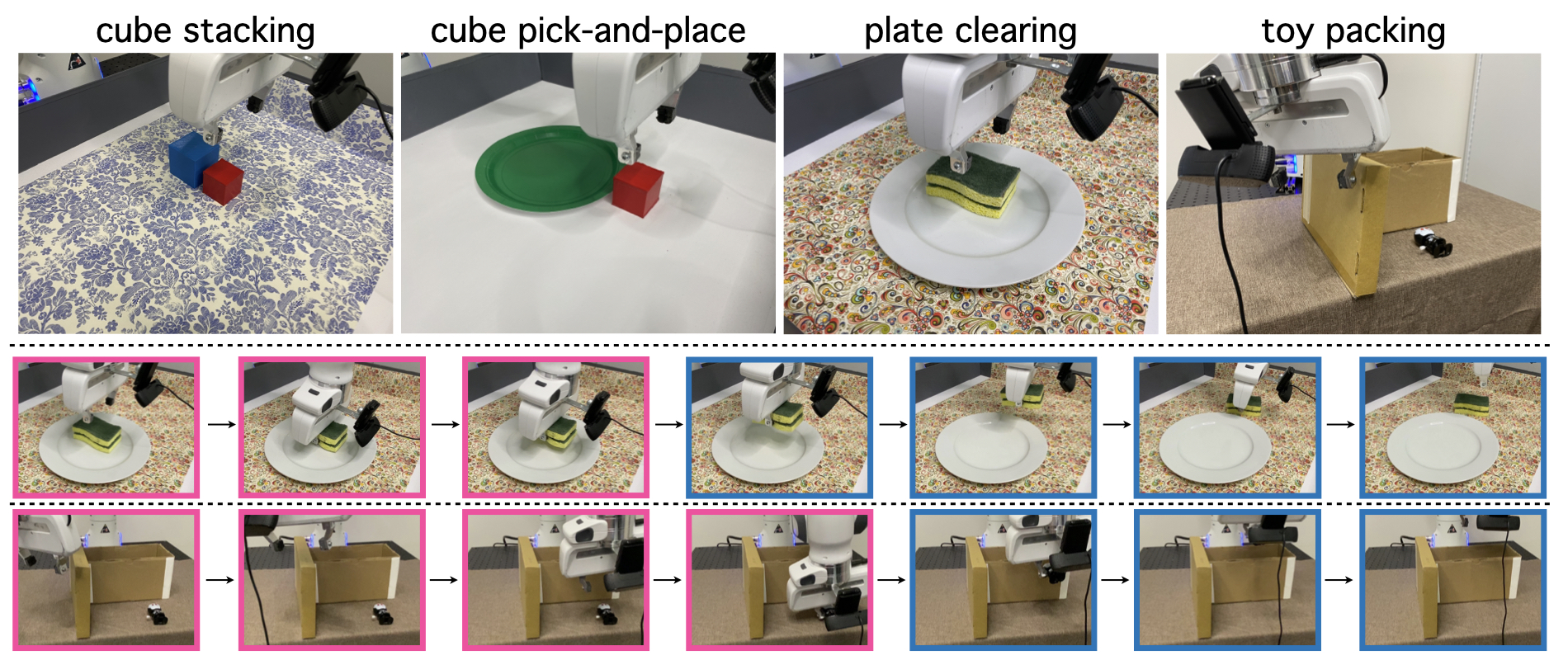}
\caption{Tasks used for task generalization experiments. Robot demonstrations perform a shorter-horizon task, such as grasping (highlighted in {\color{pink}pink}); human demonstrations either perform the full, longer-horizon task or portions of the task that are missing in the robot demonstrations (highlighted in {\color{blue}blue}). Note that while the images in {\color{blue}blue} depict a robot completing the tasks, in reality, the human demonstrations contain images of a \emph{human} completing the tasks.}
\label{fig:task-gen-tasks}
\vspace{-0.4cm}
\end{figure*}

\begin{figure}[t]
    \centering
    \includegraphics[width=\linewidth]{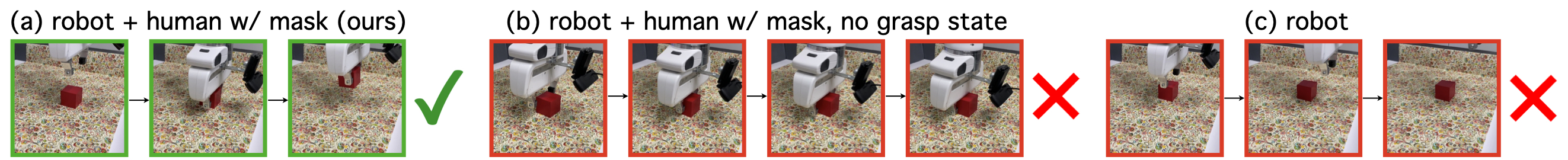}
\caption{Sample BC rollouts for cube grasping in a test environment unseen in robot demonstrations but seen in human demonstrations. Here, policies are trained on (a) both robot and human demonstrations with image masking, (b) both robot and human demonstrations with image masking but without conditioning the policy on grasp state, or (c) only robot demonstrations. In (b), the robot does not realize that it has already grasped the cube and repeatedly reattempts to do so, ultimately failing to lift the object. In (c), the robot fails to even reach the cube, as this environment background is unseen in the robot demonstrations.}
\label{fig:rollouts}
\vspace{-0.3cm}
\end{figure}

\textbf{Tasks.} The tasks we test on include stacking a red cube on top of a blue cube, picking-and-placing a red cube onto a green plate, clearing a green sponge from a plate, and packing a small black suit vampire wind-up toy into a box. See Figure \ref{fig:task-gen-tasks} for a visualization of these tasks.

\begin{table}[tb]
    \caption{\textbf{Environment generalization and task generalization results}. \textbf{Left:} BC policies are evaluated against environments not seen in the robot demonstrations. Average success rates and standard errors are computed by aggregating the results across all environments per task in Table \ref{tab:ind-env-gen-results} in Appendix \ref{app:detailed-results}. \textbf{Right:} For each experiment, robot demonstrations are collected only for the {\color{gray}\emph{gray italicized}} task, and human demonstrations are collected for the longer-horizon task written in black. Each success rate is computed over 20 test rollouts of the learned policy for the toy packing task, and 10 rollouts for the other tasks. The last row contains the average success rate across all the longer-horizon tasks. \textbf{Overall,} training with masked human demonstrations leads to significantly better environment and task generalization than other methods.}
    \vspace{4pt}
    \centering
    \resizebox{0.475\textwidth}{!}{
    \begin{tabular}{lcccc}
        \toprule
        & \multicolumn{4}{c}{environment generalization success rate (\%)} \\
        \cmidrule{2-5}
        & \multirow{2}{*}{robot} & \multirow{2}{*}{robot + play} & robot + human & \textbf{robot + human} \\
        &       &              & w/ CycleGAN   & \textbf{w/ mask (ours)} \\
        \midrule
        \multirow{2}{*}{reaching} & \multirow{2}{*}{$10.00 \pm 5.48$} & \multirow{2}{*}{$20.00 \pm 7.30$} & \multirow{2}{*}{$53.33 \pm 9.11$} & \multirow{2}{*}{$\mathbf{86.67 \pm 6.21}$} \\
        & & & & \\
        \hdashline\noalign{\vskip 0.5ex} 
        cube & \multirow{2}{*}{$0.00 \pm 0.00$} & \multirow{2}{*}{$21.67 \pm 5.32$} & \multirow{2}{*}{$36.67 \pm 6.22$} & \multirow{2}{*}{$\mathbf{51.67 \pm 6.45}$} \\
        grasping & & & & \\
        \hdashline\noalign{\vskip 0.5ex}
        plate & \multirow{2}{*}{$0.00 \pm 0.00$} & \multirow{2}{*}{$23.33 \pm 7.72$} & \multirow{2}{*}{$\mathbf{46.67 \pm 9.11}$} & \multirow{2}{*}{$\mathbf{56.67 \pm 9.05}$} \\
        clearing & & & & \\
        \hdashline\noalign{\vskip 0.5ex}
        toy & \multirow{2}{*}{$10.91 \pm 2.10$} & \multirow{2}{*}{$30.00 \pm 3.09$} & \multirow{2}{*}{$15.00 \pm 2.41$} & \multirow{2}{*}{$\mathbf{59.55 \pm 3.31}$} \\
        packing & & & & \\
        \midrule
        \multirow{2}{*}{average} & \multirow{2}{*}{$5.23 \pm 1.89$} & \multirow{2}{*}{$23.75 \pm 5.86$} & \multirow{2}{*}{$37.92 \pm 6.71$} & \multirow{2}{*}{$\mathbf{63.64 \pm 6.25}$} \\
        & & & & \\
        \bottomrule
    \end{tabular}
    }
    \hspace{0.2cm}
    \resizebox{0.49\textwidth}{!}{
    \begin{tabular}{lcccc}
        \toprule
        & \multicolumn{4}{c}{task generalization success rate (\%)} \\
        \cmidrule{2-5}
        & \multirow{2}{*}{robot} & \multirow{2}{*}{robot + play} & robot + human & \textbf{robot + human} \\
        &       &              & w/ CycleGAN   & \textbf{w/ mask (ours)} \\
        \midrule
        {\color{gray}\emph{cube grasping}} & {\color{gray}$90$} & {\color{gray}$90$} & {\color{gray}$80$} & {\color{gray}$90$} \\
        cube stacking & $0$ & $10$ & $0$ & $\mathbf{40}$ \\
        \hdashline\noalign{\vskip 0.5ex}
        {\color{gray}\emph{cube grasping}} & {\color{gray}$90$} & {\color{gray}$100$} & {\color{gray}$90$} & {\color{gray}$90$} \\
        cube pick-and-place & $0$ & $0$ & $20$ & $\mathbf{80}$ \\
        \hdashline\noalign{\vskip 0.5ex}
        {\color{gray}\emph{sponge grasping}} & {\color{gray}$100$} & {\color{gray}$100$} & {\color{gray}$100$} & {\color{gray}$100$} \\
        plate clearing & $0$ & $10$ & $30$ & $\mathbf{70}$ \\
        \hdashline\noalign{\vskip 0.5ex}
        {\color{gray}\emph{toy grasping}} & {\color{gray}$95$} & {\color{gray}$75$} & {\color{gray}$100$} & {\color{gray}$100$} \\
        toy packing & $0$ & $15$ & $0$ & $\mathbf{45}$ \\
        \midrule
        \multirow{2}{*}{average} & \multirow{2}{*}{$0$} & \multirow{2}{*}{$8.75$} & \multirow{2}{*}{$12.5$} & \multirow{2}{*}{$\mathbf{58.75}$} \\
        & & & & \\
        \bottomrule
    \end{tabular}
    }
    \label{tab:env-gen-and-task-gen-results}
    \vspace{-0.4cm}
\end{table}

\textbf{Datasets.} As in Section \ref{sec:env-gen-experiments}, we collect robot demonstrations, human demonstrations, and shared robot play data. Robot demonstrations perform a simple, short-horizon task (e.g., cube grasping), and human demonstrations perform one of the more difficult, longer-horizon tasks above (e.g., cube stacking). Appendix \ref{app:datasets} gives full details on all datasets used in the experiments.

\textbf{Methods.} We evaluate the task generalization of the same four methods discussed in Section \ref{sec:env-gen-experiments}.

\textbf{Results.} As shown in Table \ref{tab:env-gen-and-task-gen-results}, training the policy on the eye-in-hand human video demonstrations with image masking substantially improves task generalization compared to using robot data alone. Intuitively, a policy trained on robot demonstrations that never perform the desired multi-stage task is incapable of performing the task at test time. A policy that is also trained on robot play data can occasionally execute the desired task since the play dataset contains a collection of behaviors, some of which can be useful for solving the task. However, as the play dataset is task-agnostic, BC often struggles to learn one coherent sequence of actions for solving a specific multi-stage task. Lastly, a policy trained on human demonstrations translated to the robot domain via CycleGAN can generalize to new tasks, but it performs worse than simply using our proposed image masking scheme due to the aforementioned errors in image translation (Figure~\ref{fig:cyclegan-examples} in Appendix~\ref{app:cyclegan-analysis}). See further qualitative analyses of individual methods and videos of learned policies on our \href{https://giving-robots-a-hand.github.io/}{project website}.

\vspace{-0.2cm}
\subsection{Ablation Experiments}
\label{sec:ablation-experiments}
\vspace{-0.2cm}

\textbf{Training with unmasked images.} We remove the image masking entirely to assess whether it is an important component of our framework. Given unmasked robot play data where the end-effector is now visible, we train an inverse model to predict the dynamics and use the model to infer action labels for unmasked human demonstrations, regardless of the domain shift caused by visual differences between the human hand and the robot gripper. (Note that we do not use CycleGAN image translation here.) We train a BC policy on unmasked versions of the robot and human demonstrations used in our previous experiments, and compare this to our original method.

\begin{wraptable}{r}{6.5cm}
    \caption{\textbf{Ablation experiments results.}
    We observe that removing either the image masking or grasp state conditioning generally leads to greatly reduced success rates, validating their important contributions to the final generalization performance. Success rates and standard errors are computed by aggregating the finer-grained results in Table \ref{tab:ind-ablation-results}.}
    \label{tab:ablation-results}
    \centering
    \resizebox{0.47\textwidth}{!}{
    \begin{tabular}{lc}
        \toprule
        & success rate (\%) \\
        \midrule
        robot + human w/ mask (ours) & $54.29 \pm 5.95$ \\
        robot + human, no mask & $24.29 \pm 5.13$ \\
        robot + human w/ mask, no grasp state & $28.57 \pm 5.40$ \\
        \bottomrule
    \end{tabular}
    }
\end{wraptable}

\textbf{Behavioral cloning without conditioning on grasp state.} In a separate ablation, we modify the BC policy such that it is no longer conditioned on the binary (open/close) grasp state. We reuse the robot and human demonstrations from the previous experiments and simply train a new BC policy without grasp state inputs. Note that we are still masking images here, as in our original framework.

\textbf{Results.} As shown in Table \ref{tab:ablation-results} (detailed results are shown in Table \ref{tab:ind-ablation-results}), removing either image masking or grasp state reduces overall performance. Qualitatively, the policy often fails to even reach the target object in several cases when using unmasked images; we attribute this to the distribution shift between human and robot observations.
Without conditioning on grasp state, a common failure mode we observe is repeatedly attempting to grasp an object rather than lifting it, as the robot does not know that it has already secured the object (an illustration of this behavior is shown in Figure \ref{fig:rollouts}(b)). Overall, both components here are important to successfully leverage eye-in-hand human video demonstrations.

\vspace{-0.2cm}
\section{Conclusion}
\label{sec:conclusion}
\vspace{-0.2cm}

This work presents a novel yet simple framework for leveraging diverse eye-in-hand human video demonstrations and displays its potential to enhance the generalization of vision-based manipulators. We utilize eye-in-hand cameras and image masking to largely close the domain gap between human and robot data and bypass explicit domain adaptation entirely. Our framework enables an imitation learning policy to generalize to new environments and new tasks unseen in the robot demonstrations.

\textbf{Limitations and future work.}
Our image masking scheme may not be as effective if the target object is so minuscule ($<1.5$ cm long on each side) that it is not visible in the unmasked portion of the image, because it may be difficult for the inverse model to infer actions that manipulate the object due to insufficient visual cues. Optimizing the camera angle so that a smaller portion of the image can be masked could mitigate this issue. Additionally, our method involves collecting a robot play dataset to train the inverse model. While this process is inexpensive (details discussed in Appendix~\ref{app:play-data-details}), in the future we hope to automate play data collection nonetheless, e.g., by training a BC policy on a small play dataset and sampling actions during inference to encourage exploration (as in \citep{dinyari2020learning}).

\clearpage
\acknowledgments{We thank Alexander Khazatsky, Tony Zhao, Suraj Nair, Kaylee Burns, Maximilian Du, and other members of the Stanford IRIS Lab for insightful discussions and helpful feedback. Moo Jin Kim gratefully acknowledges the financial support of the Siebel Scholarship. This work was supported by ONR grant N00014-21-1-2685.}


\bibliography{refs}  

\clearpage
\appendix
\section{Appendix}

\subsection{Model Architectures}
\label{app:model-architectures}

In this section, we discuss model architecture details. We implement and train all models using PyTorch \citep{paszke2019pytorch}.

\subsubsection{Inverse Dynamics Model Architecture}
\label{app:inverse-model-arch}

The inverse dynamics model is a convolutional neural network with $4$ convolutional layers followed by $2$ feedforward layers. Each convolutional and feedforward layer is followed by a batch normalization layer and a ReLU activation layer. For every convolutional layer, the number of convolutional filters is 128, kernel size is $3$, stride is $1$ (except for the first layer, whose stride is $2$), and padding is $0$. The latent embedding size of the second feedforward layer is $200$. We use early fusion, i.e., two consecutive image observations are concatenated channel-wise and then fed into the first convolutional layer. The full network outputs an action prediction that takes the agent from one observation to the next timestep's observation, where the action is 3-DoF or 6-DoF with an additional binary gripper action. (Recall that we use 6-DoF position and orientation control for the toy packing task, and 3-DoF position control for the other tasks.)

We train every inverse model with random shifts data augmentation. For every pair of $100 \times 100$ image observations, we pad each side by $4$ pixels and randomly crop a $100 \times 100$ region out of the result. The same augmentation is applied to both images in a given pair so as to not perturb the original dynamics captured in the images. We only apply this augmentation with $80\%$ probability, as we found that the resulting model is as accurate as one trained with $100\%$ probability, yet it trains faster because it does not need to compute the augmentation $20$ percent of the time.

\subsubsection{Behavioral Cloning Policy Network Architecture}
\label{app:bc-policy-arch}

The BC policy network consists of an image encoder with mostly the same architecture as the inverse model, except that the number of convolutional filters per layer is $32$, and the hidden size of the second feedforward layer is $50$. Unlike the inverse model, the policy network acts on one image at a time rather than a pair. After the image encoder portion, the policy network consists of an additional two feedforward layers (with a latent dimensionality of $64$) representing the policy head. Further, the policy is conditioned on a $1$-dimensional grasp state variable as described in Section \ref{sec:bc-with-human-data}; this variable is concatenated with the $50$-dimensional latent embedding output by the second feedforward layer of the image encoder, and the resulting $51$-dimensional embedding is passed on to the policy head, which outputs an action prediction that best imitates the expert demonstrator's action given some input observation.

As with the inverse model, we apply random shifts data augmentation while training the BC policy.

\subsection{Tasks}
\label{app:task-details}

In this section, we discuss the tasks introduced in Section \ref{sec:experiments} in more detail. All tasks involve 3-DoF position control (and 1-DoF binary gripper control), except for the toy packing tasks, which involve 6-DoF position/orientation control.

\subsubsection{Environment Generalization Tasks}
The tasks used for the environment generalization experiments include the following:

\begin{itemize}
  \item \textbf{reaching}: The goal is to reach the end-effector towards the red cube. The environment contains just the red cube; or the red cube and a blue cube distractor; or the red cube and a green sponge distractor; or all three objects. The initial positions of the objects are randomized within a $50 \times 50$ cm section of the environment.
  \item \textbf{cube grasping}: The goal is to grasp the red cube and lift it off the ground. The cube is the only object in the environment. The environment background can be one of seven: plain white background, rainbow floral texture, green floral texture, blue floral texture, orange plate, green plate, or blue plate. The initial position of the cube is randomized within a $30 \times 30$ cm section of the environment.
  \item \textbf{plate clearing}: The goal is to grasp a target object resting on a plate, lift it up, and transfer it to a location off to the right of the plate. The target object is either a green sponge, yellow sponge, blue towel, or pink towel. The initial position of the target object is randomized within a $20 \times 20$ cm section of the plate.
  \item \textbf{toy packing}: The goal is to maneuver around a wall, reach towards the toy, grasp it, lift it up, move over to the open box, and release the toy into the box. In the initial environment setting, the end-effector is positioned behind a thin cardboard box (which we call the ``wall'') such that the toy is fully occluded in the eye-in-hand image observations at the beginning of the episode. There are twelve types of toys that our policies are evaluated against (see Figure \ref{fig:env-gen-tasks} for pictures of the toys): 
  black suit vampire toy,
  white mummy toy,
  orange body jack-o'-lantern toy,
  red cape vampire toy,
  purple body green zombie toy,
  crazy witch toy,
  green body jack-o'-lantern toy,
  purple body jack-o'-lantern,
  red dentures w/ USA hat toy,
  green Christmas tree toy,
  Santa Claus toy, and
  brown reindeer toy.
  The initial positions of the toy, end-effector, and two boxes are all randomized within a $40 \times 10$ cm section of the table, and the toy's position relative to the boxes is also randomized within a $15 \times 10$ cm section in front of the open box. In addition, the end-effector is initially angled towards the wall, as opposed to being oriented top-down as in the other tasks (see Figure \ref{fig:env-gen-tasks} for an illustration). Therefore, 6-DoF control is necessary for solving the task.
\end{itemize}

\subsubsection{Task Generalization Tasks}
We now describe the tasks used in the task generalization experiments:

\begin{itemize}
  \item \textbf{cube stacking}: The goal is to grasp the red cube, lift it up, stack it on top of the blue cube, and release the red cube. The initial positions of the cubes are fixed relative to each other, but vary relative to the blue floral texture background within a $20 \times 20$ section of the environment. Robot demonstrations perform cube grasping, while human demonstrations perform full cube stacking or portions of the task that follow the grasp.
  \item \textbf{cube pick-and-place}: The goal is to grasp the red cube, lift it up, move it over to the green plate, and release it onto the plate. The initial positions of the cube and plate are fixed relative to each other, but vary relative to the plain white background within a $20 \times 20$ section of the environment. Robot demonstrations perform cube grasping, while human demonstrations perform full cube pick-and-place or portions of the task that follow the grasp.
  \item \textbf{plate clearing}: The goal is the same as described earlier for the plate clearing environment generalization task. However, here we only manipulate one target object: the green sponge. The initial position of the sponge is randomized within a $20 \times 20$ cm section of the plate. Robot demonstrations perform sponge grasping, while human demonstrations perform full plate clearing or portions of the task that follow the grasp.
  \item \textbf{toy packing}: The goal is the same as described earlier for the toy packing environment generalization task. However, here we only manipulate one target object: the black suit vampire toy. As before, the initial position of the toy, end-effector, and two boxes are randomized within a $40 \times 10$ cm section of the table, and the toy's position relative to the boxes is also randomized within a $15 \times 10$ cm section in front of the open box. Robot demonstrations perform toy grasping, while human demonstrations perform full toy packing or portions of the task that follow the grasp.
\end{itemize}

Please see our  \href{https://giving-robots-a-hand.github.io/}{project website} for further details and visualizations of data collected in these tasks and environments (expand the page using the button at the very bottom).

\subsection{Datasets}
\label{app:datasets}

\subsubsection{Robot Play Datasets}
\label{app:play-data-details}
\textbf{How play data is collected.} We gather play data in a similar manner as \citet{lynch2020learning}: a human teleoperator controlling a Franka Emika Panda robot arm executes a diverse repertoire of behaviors in an environment, exploring the observation and action spaces while interacting with objects in the scene. For example, in an environment containing two cubes, the teleoperator may wave the robotic end-effector around, reach towards a cube, grasp and lift up a cube, release and drop the cube, stack one cube on top of the other, and so on. The continuous sequences of observations captured by the eye-in-hand camera and the actions commanded by the teleoperator are logged and stored into a replay buffer $\mathcal{D}_{\text{play}}^r$ for inverse model training. See the subsections below and the \href{https://giving-robots-a-hand.github.io/}{project website} for examples of play datasets.

\textbf{Why play data is easy to collect.} The key advantage of using play data is that it is easy to collect meaningful interaction data in large quantities \citep{lynch2020learning} due to the following:
\begin{itemize}
  \item There is no need to frequently reset the manipulator and objects to some initial state (which is typically necessary when collecting expert demonstrations).
  \item There is no notion of maximum episode length or time limit (allowing a teleoperator to execute a variety of behaviors in a single contiguous stretch of time, pausing only when desired).
  \item The teleoperator's knowledge of object affordances leads to interesting interactions with objects (as opposed to a script that executes purely random actions, which leads to slower exploration of the interaction space unless the data collection process is manually biased towards more meaningful interactions, as in \citep{nair2017combining}).
  \item The play behaviors do not have to solve any particular task (which makes it easier to collect play data than expert task-specific demonstrations).
\end{itemize}
As a result, we can quickly collect a play dataset for a given environment, or set of environments, that is sufficient for training the inverse dynamics model. In addition, a single play dataset could in principle be used to develop an inverse model that is reused for many different downstream tasks, effectively amortizing the cost of collecting it.

\textbf{Details on collected play datasets.} We collect four robot play datasets and train four corresponding inverse models. Each inverse model is shared across one environment generalization experiment and one task generalization experiment. We discuss the details of each play dataset below:

\begin{itemize}
  \item \textbf{reaching and cube stacking dataset}: We collect $20{,}000$ timesteps of play data at $5$ Hz (approximately $67$ minutes) in an environment with a blue floral background and three objects: a red cube, a blue cube, and a green sponge. The play data behaviors include waving the end-effector around, reaching towards each object, grasping and lifting up each object, releasing and dropping an object, stacking an object on top of another, and so on. This play dataset is shared for the reaching environment generalization and cube stacking task generalization tasks.
  \item \textbf{cube grasping and cube pick-and-place dataset}: We collect $52{,}400$ steps of play data at $5$ Hz (approximately $171$ minutes) in multiple environments containing a red cube, each having a different background that the red cube rests on: plain white background, rainbow floral texture, green floral texture, blue floral texture, orange plate, green plate, or blue plate. The play data behaviors include waving the end-effector around, reaching towards the cube, grasping and lifting up the cube, releasing and dropping the cube, and so on. This play dataset is shared for the cube grasping environment generalization and cube pick-and-place task generalization tasks.
  \item \textbf{plate clearing environment generalization and task generalization dataset}: We collect $20{,}000$ steps of play data at $5$ Hz (approximately $67$ minutes) in multiple environment configurations, each containing a different target object: green sponge, yellow sponge, blue towel, and pink towel. The play data behaviors include waving the end-effector around, reaching toward the objects, grasping and lifting up the objects, releasing and dropping the objects, and so on. This play dataset is shared for both plate clearing environment generalization and task generalization experiments.
  \item \textbf{toy packing environment generalization and task generalization dataset}: We collect $10{,}000$ steps of play data at $4$ Hz (approximately $42$ minutes) in multiple environment configurations, each containing a different target toy. The toys included in this dataset are the following:
  white mummy toy,
  orange body jack-o'-lantern toy,
  green body jack-o'-lantern toy,
  crazy witch toy,
  purple body green zombie toy,
  purple body jack-o'-lantern,
  black suit vampire toy,
  red dentures w/ USA hat toy,
  red cape vampire toy,
  pirate bomb toy,
  green witch w/ broomstick toy,
  purple dentures w/ eyes toy,
  eyeball toy,
  skull toy, and
  X-ray skeleton toy.
  The play data behaviors include waving the end-effector around, reaching around the wall and towards the boxes, reaching towards the toys, grasping and lifting up the toys, releasing and dropping the toys, and so on. This play dataset is shared for both toy packing environment generalization and task generalization experiments.
\end{itemize}

Please see our  \href{https://giving-robots-a-hand.github.io/}{project website} for visualizations of these play datasets (expand the page using the button at the very end).

\subsubsection{Expert Demonstration Datasets}
\label{app:expert-demo-details}

In each environment generalization or task generalization experiment, we collect a set of expert robot demonstrations and a set of expert human demonstrations. Below we discuss details of the datasets collected for each experiment. Please refer to Figure \ref{fig:env-gen-tasks} and Figure \ref{fig:task-gen-tasks} for a visualization of the distribution of environments or tasks that the robot and human datasets are each collected from. All demonstrations are collected at 5 Hz (or 4 Hz for the toy packing tasks), as is done while collecting the play datasets.

\begin{itemize}
  \item \textbf{reaching (environment generalization)}: We collect $60$ robot demonstrations with no distractor objects and $100$ human demonstrations with both the blue cube and green sponge as distractors.
  \item \textbf{cube grasping (environment generalization)}: We collect $100$ robot demonstrations only in an environment with a plain white background and $20$ human demonstrations from each of the following environment backgrounds: rainbow floral texture, green floral texture, blue floral texture, orange plate, green plate, and blue plate.
  \item \textbf{plate clearing (environment generalization)}: We collect $30$ robot demonstrations with just the green sponge as a target object and $20$ human demonstrations with each of the following target objects: yellow sponge, blue towel, and pink towel.
  \item \textbf{toy packing (environment generalization)}: We collect $100$ robot demonstrations with just the black suit vampire toy and $20$ human demonstrations with each of the following toys: white mummy toy, orange body jack-o'-lantern toy, red cape vampire toy, purple body green zombie toy, and crazy witch toy.
  \item \textbf{cube stacking (task generalization)}: We collect $25$ robot demonstrations and $130$ human demonstrations. The robot demonstrations perform red cube grasping; the human demonstrations perform cube stacking (stack red cube onto blue cube) or portions of the task that follow the grasp. For this task, a majority of the human demonstrations do the latter and are thus able to be collected very quickly.
  \item \textbf{cube pick-and-place (task generalization)}: We collect $20$ robot demonstrations and $70$ human demonstrations. The robot demonstrations perform cube grasping; the human demonstrations perform cube pick-and-place (place cube onto plate)  or portions of the task that follow the grasp.
  \item \textbf{plate clearing (task generalization)}: We collect $40$ robot demonstrations and $25$ human demonstrations. The robot demonstrations perform sponge grasping; the human demonstrations perform plate clearing (remove green sponge off of plate) or portions of the task that follow the grasp.
  \item \textbf{toy packing (task generalization)}: We collect $100$ robot demonstrations and $100$ human demonstrations. The robot demonstrations perform toy grasping; the human demonstrations perform toy packing (lift the toy and drop it into the box) or portions of the task that follow the grasp.
\end{itemize}

Please see our  \href{https://giving-robots-a-hand.github.io/}{project website} for visualizations of these expert demonstration datasets (expand the page using the button at the very end).

\subsection{CycleGAN Analysis}
\label{app:cyclegan-analysis}

\begin{figure}[h]
    \centering
    \includegraphics[width=0.98\linewidth]{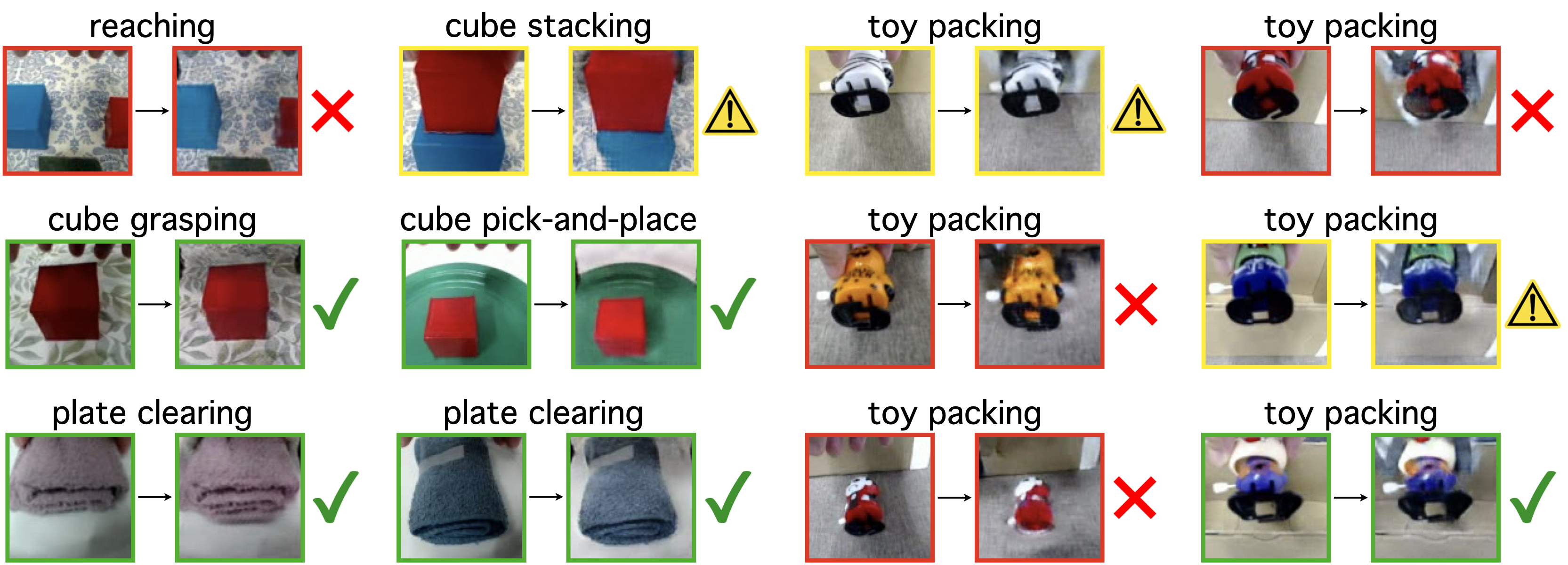}
    \caption{Sample human-to-robot image translations in various tasks via CycleGAN. Explicit image translation provides an alternative approach for leveraging human video demonstrations. In some cases, the human embodiment is successfully translated into the robot domain (highlighted in {\color{dark-green}green}), even in cases where there are five human fingers and two robot fingers (e.g., in the second row, first two columns of images). However, in other cases, noticeable visual artifacts exist after the translation (highlighted in {\color{yellow}yellow} in moderate cases, or {\color{red}red} in severe cases).}
    \label{fig:cyclegan-examples}
\end{figure}

In Figure \ref{fig:cyclegan-examples}, we show sample human-to-robot image translations output by CycleGAN \citep{zhu2017unpaired}. The translations are successful in some cases but noisy in others. Noisy translations hinder final BC policy performance, resulting in lower performance than simple image masking.

\newpage
\subsection{Inverse Dynamics Model Analysis}
\label{app:inverse-model-analysis}

\subsubsection{Sample Inverse Dynamics Model Action Predictions}

\begin{figure}[h]
    \centering
    \includegraphics[width=0.98\linewidth]{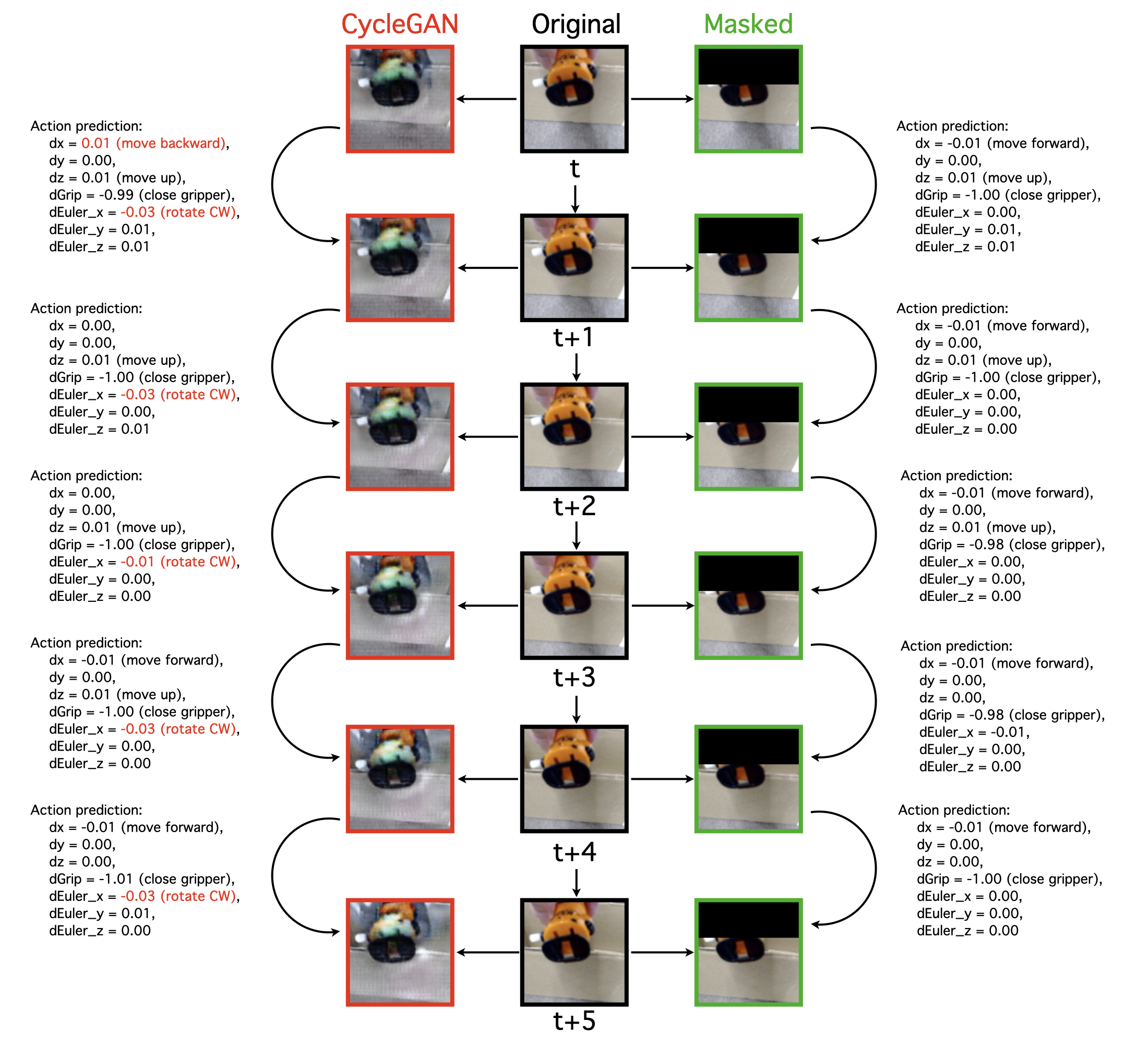}
    \caption{Sample inverse dynamics model action predictions.}
    \label{fig:inverse-model-predictions}
\end{figure}

In Figure \ref{fig:inverse-model-predictions}, we show sample outputs from the inverse dynamics model when labeling human video demonstrations with actions for the toy packing task. The middle column contains the original human images. The leftmost column contains human images translated to the robot domain via CycleGAN. The rightmost column contains human images masked according to our proposed image masking scheme. We highlight major mistakes made by the inverse model in {\color{red}red}. Due to noise in the CycleGAN translations, we see that there exists a significant nonzero rotation component in several of the action predictions in the leftmost column, which causes the robot gripper to rotate excessively in some cases (we show such behavior in the videos for \textbf{robot + human w/ CycleGAN} on our \href{https://giving-robots-a-hand.github.io/}{project website}). In contrast, we avoid such issues using our image masking method.

\subsubsection{Validating Inverse Dynamics Model Accuracy}

Learning an accurate inverse dynamics model is not unusually challenging given that we leverage eye-in-hand camera observations in this work. Suppose we have a tuple $(o, a, o^\prime)$, where $o$ represents the current image observation, $a$ represents the current action, and $o^\prime$ represents the next image observation. Recall that the inverse dynamics modeling problem is to predict the action $a$ giving rise to the change in observations. Predicting the action is fairly intuitive in our framework: for example, if an object in the eye-in-hand camera view is moving to the right, we can infer that the hand or gripper is moving to the left. The inverse model can use any visual cue in the scene as a reference point while learning to predict the dynamics.

Regardless of the perceived difficulty, there are several ways we can validate the behavior of a learned inverse model. Quantitatively, we can check the performance of the inverse dynamics model on a validation set (e.g., held-out robot play data). Qualitatively, we can check whether the predicted actions for a human video are sensible. For instance, we can verify that the action predictions are smooth and not noisy, e.g., by outputting a chunk of observation-action pairs and observing a coherent action trajectory over a continuous stretch of time.

Lastly, we note that, if desired, one could avoid collecting play data and training an inverse model by inferring actions via visual odometry or Structure-from-Motion pose estimation methods.

\newpage
\subsection{Detailed Experimental Results}
\label{app:detailed-results}

Table \ref{tab:ind-env-gen-results} contains the full experimental results that were aggregated to produce Table \ref{tab:env-gen-and-task-gen-results} in Section \ref{sec:experiments}. All success rates are evaluated over 10 trials for all tasks (except for the toy packing task, in which we use 20 trials to compute each success rate). Initial object positions are distributed according to the configurations described in Appendix \ref{app:task-details}. Please see our \href{https://giving-robots-a-hand.github.io/}{project website} for videos of the learned policies.

In addition, Table \ref{tab:ind-ablation-results} contains detailed ablation experiment results that were summarized to produce Table \ref{tab:ablation-results} in Section \ref{sec:ablation-experiments}.

\definecolor{gray}{rgb}{.5,.5,.5}
\vspace{0.5cm}
\begin{table*}[h]
    \caption{\textbf{Full environment generalization experiments results (aggregate results in Table \ref{tab:env-gen-and-task-gen-results}).} BC policies are trained on only robot demonstrations, robot demonstrations and robot play data, robot demonstrations and CycleGAN-translated human demonstrations, or robot and human demonstrations with image masking. For each task, the robot demonstrations are collected only in the {\color{gray}\emph{gray italicized}} environment configuration in the second column; play data and human demonstrations are collected in the configurations below the dotted lines. Thus, the non-italicized environment configurations are out-of-distribution with respect to the robot demonstrations, and these are the configurations we focus on. Further, the environment configurations highlighted in {\color{dark-green}green} are not seen in the human demonstrations but are seen in the robot play data, while the configurations highlighted in {\color{yellow}yellow} are not seen at all in either human or robot play/demonstration data (i.e., these are fully held out). Overall, leveraging human demonstrations leads to significantly greater environment generalization than using robot demonstrations alone, and performs better than training on robot play data and CycleGAN-translated human data as well. Each success rate is computed over 20 test rollouts of the learned policy for the toy packing task, and 10 rollouts for the rest.}
    \label{tab:ind-env-gen-results}
    \centering
    \resizebox{\textwidth}{!}{
    \begin{tabular}{llcccc}
        \toprule
        & & \multicolumn{4}{c}{success rate (\%)} \\
        \cmidrule{3-6}
        \multirow{2}{*}{task} & \multirow{2}{*}{environment configuration} & \multirow{2}{*}{robot} & \multirow{2}{*}{robot + play} & robot + human & \textbf{robot + human} \\
        & &       &              & w/ CycleGAN   & \textbf{w/ mask (ours)} \\
        \midrule
        \multirow{4}{*}{reaching} & {\color{gray}\emph{no distractors (only red cube)}} & {\color{gray}$90$} & {\color{gray}$90$} & {\color{gray}$80$} & {\color{gray}$90$} \\
        \cdashline{2-6}
        & + blue cube distractor & $20$ & $20$ & $70$ & $\mathbf{90}$ \\
        & + green sponge distractor & $10$ & $20$ & $30$ & $\mathbf{90}$ \\
        & + blue cube, green sponge distractors & $0$ & $20$ & $60$ & $\mathbf{80}$ \\
        \midrule
        \multirow{7}{*}{cube grasping}
        & {\color{gray}\emph{white background}} & {\color{gray}$90$} & {\color{gray}$90$} & {\color{gray}$100$} & {\color{gray}$90$} \\
        \cdashline{2-6}
        & rainbow floral background & $0$ & $30$ & $10$ & $\mathbf{80}$ \\
        & green floral background & $0$ & $20$ & $50$ & $\mathbf{60}$ \\
        & blue floral background & $0$ & $30$ & $\mathbf{70}$ & $60$ \\
        & cube on orange plate & $0$ & $20$ & $\mathbf{40}$ & $\mathbf{40}$ \\
        & cube on green plate & $0$ & $10$ & $0$ & $\mathbf{20}$ \\
        & cube on blue plate & $0$ & $20$ & $\mathbf{50}$ & $\mathbf{50}$ \\
        \midrule
        \multirow{4}{*}{plate clearing} & {\color{gray}\emph{green sponge on plate}} & {\color{gray}$60$} & {\color{gray}$70$} & {\color{gray}$80$} & {\color{gray}$70$} \\
        \cdashline{2-6}
        & yellow sponge on plate & $0$ & $30$ & $\mathbf{50}$ & $40$ \\
        & blue towel on plate & $0$ & $10$ & $20$ & $\mathbf{70}$ \\
        & pink towel on plate & $0$ & $30$ & $\mathbf{70}$ & $60$ \\
        \midrule
        \multirow{12}{*}{toy packing}
        & {\color{gray}\emph{black suit vampire toy}} & {\color{gray}$85$} & {\color{gray}$70$} & {\color{gray}$80$} & {\color{gray}$85$} \\
        \cdashline{2-6}
        & white mummy toy & $40$ & $45$ & $40$ & $\mathbf{60}$ \\
        & orange body jack-o'-lantern toy & $0$ & $20$ & $15$ & $\mathbf{45}$ \\
        & red cape vampire toy & $0$ & $20$ & $0$ & $\mathbf{65}$ \\
        & purple body green zombie toy & $35$ & $45$ & $35$ & $\mathbf{75}$ \\
        & crazy witch toy & $30$ & $40$ & $30$ & $\mathbf{75}$ \\
        & {\color{dark-green}green body jack-o'-lantern w/ hat toy} & $0$ & $25$ & $0$ & $\mathbf{55}$ \\
        & {\color{dark-green}purple body jack-o'-lantern toy} & $0$ & $30$ & $0$ & $\mathbf{75}$ \\
        & {\color{dark-green}red dentures w/ USA hat toy} & $0$ & $20$ & $5$ & $\mathbf{30}$ \\
        & {\color{yellow}green Christmas tree toy} & $5$ & $15$ & $5$ & $\mathbf{35}$ \\
        & {\color{yellow}Santa Claus toy} & $0$ & $35$ & $25$ & $\mathbf{65}$ \\
        & {\color{yellow}brown reindeer toy} & $10$ & $35$ & $10$ & $\mathbf{75}$ \\
        \bottomrule
    \end{tabular}
    }
\end{table*}

\newpage
\begin{table*}[t]
    \caption{\textbf{Full ablation experiments results.} We test policies on one representative task from the environment generalization setting (cube grasping) and another from the task generalization setting (plate clearing). The policy trained on unmasked images fails drastically on the last three environment configurations as it never reaches the cube. The policy that is not conditioned on the grasp state often encounters a failure mode in which it repeatedly reattempts to grasp an object even though it has already grasped it. Such a failure mode occurs because the robot cannot see the end-effector. Each success rate is computed over 10 test rollouts of the BC policy.}
    \label{tab:ind-ablation-results}
    \centering
    \resizebox{0.8\textwidth}{!}{
    \begin{tabular}{llcccc}
        \toprule
        \multicolumn{5}{c}{environment generalization} \\
        \toprule
        & & \multicolumn{4}{c}{success rate (\%)} \\
        \cline{3-6}
        task & environment configuration & original & no image masking & no grasp state \\
        \midrule
        \multirow{3}{*}{cube grasping}
        & \emph{white background} & $90$ & $40$ & $90$ \\
        \cdashline{2-5}
        & rainbow floral background & $\mathbf{80}$ & $60$ & $50$ \\
        & green floral background & $\mathbf{60}$ & $40$ & $40$ \\
        & blue floral background & $\mathbf{60}$ & $40$ & $20$ \\
        & cube on orange plate & $\mathbf{40}$ & $0$ & $0$ \\
        & cube on green plate & $\mathbf{20}$ & $0$ & $10$ \\
        & cube on blue plate & $\mathbf{50}$ & $0$ & $10$ \\
        \bottomrule
        \toprule
        \multicolumn{5}{c}{task generalization} \\
        \toprule
        & & \multicolumn{4}{c}{success rate (\%)} \\
        \cline{3-6}
        experiment & task & original & no image masking & no grasp state \\
        \midrule
        \multirow{2}{*}{$1$} & \emph{sponge grasping} & $100$ & $90$ & $90$ \\
        \cdashline{2-5}
        & clearing sponge from plate & $\mathbf{70}$ & $30$ & $\mathbf{70}$ \\
        \bottomrule
    \end{tabular}
    }
\end{table*}

\end{document}